\documentclass[11pt, twocolumn]{article}

\usepackage[utf8]{inputenc}
\usepackage[T1]{fontenc}
\usepackage{amsmath,amsfonts,amssymb}
\usepackage{graphicx}
\usepackage{hyperref}
\usepackage{cite}
\usepackage{url}
\usepackage[authoryear]{natbib}

\usepackage[top=1in, bottom=1in, left=0.75in, right=0.75in, columnsep=0.25in]{geometry}

\usepackage{setspace}
\usepackage{fancyhdr}
\usepackage{float}
\usepackage{subcaption}
\usepackage{booktabs}
\usepackage{algorithm}
\usepackage{algorithmic}

\usepackage{stfloats} 

\newcommand{\GDMP}{$\mathcal{G}$-DMP}

\begin{document}

\title{A Survey on Imitation Learning for Contact-Rich Tasks in Robotics}

\date{\today}

\author{
Toshiaki Tsuji$^{1,*}$ \and
Yasuhiro Kato$^{2,*}$ \and
Gokhan Solak$^{3,*}$ \and
Heng Zhang$^{3,4,*}$ \and
Tadej Petrič$^{5,*}$ \and
Francesco Nori$^{6,*}$ \and
Arash Ajoudani$^{3,*}$
}

\twocolumn[
\begin{@twocolumnfalse}
\maketitle

\begin{center}
\footnotesize
$^1$Graduate School of Science and Engineering, Saitama University, Saitama, Japan\\
$^2$Center for Education and Research in Information Science and Technology, University of Tokyo, Tokyo, Japan\\
$^3$Human-Robot Interfaces and Interaction Lab, Istituto Italiano di Tecnologia, Genova, Italy\\
$^4$Ph.D. program of national interest in Robotics and Intelligent Machines (DRIM) and Università di Genova, Genoa, Italy\\
$^5$Department of Automatics, Biocybernetics and Robotics, Jožef Stefan Institute, Ljubljana, Slovenia\\
$^6$Robotics lab, Google Deepmind\\
$^*$All authors contributed equally and majorly to this review paper.
\end{center}
\vspace{0.3cm}

\begin{abstract}
This paper comprehensively surveys research trends in imitation learning for contact-rich robotic tasks. Contact-rich tasks, which require complex physical interactions with the environment, represent a central challenge in robotics due to their nonlinear dynamics and sensitivity to small positional deviations. The paper examines demonstration collection methodologies, including teaching methods and sensory modalities crucial for capturing subtle interaction dynamics. We then analyze imitation learning approaches, highlighting their applications to contact-rich manipulation. Recent advances in multimodal learning and foundation models have significantly enhanced performance in complex contact tasks across industrial, household, and healthcare domains. Through systematic organization of current research and identification of challenges, this survey provides a foundation for future advancements in contact-rich robotic manipulation.\\

\end{abstract}
\vspace{0.5cm}
\end{@twocolumnfalse}
]

\section{Introduction}
Robots are intelligent systems that bring physical effects to real environments. 
Many basic tasks are contact-rich tasks involving multiple contact states between robots 
and their environment, and developing these capabilities is one of the core challenges in robotics. 
Meanwhile, understanding ``everyday physics'' has long been known to be extremely 
difficult~\citep{arimoto1999robotics}. 
Our daily tasks involve complex interactions of diverse physical phenomena such as friction, 
elasticity, plastic deformation, and fracture, which exhibit nonlinear and unpredictable behaviors, 
making the advancement of contact-rich tasks both an old and new research challenge.

Approaches to contact-rich tasks can be broadly categorized into model-based and model-free methods, with model-based approaches being widely studied at the practical level~\citep{xu2019compare}. However, as research in everyday physics indicates, contact-rich tasks require highly nonlinear models since slight positional deviations can cause significant behavioral changes. As these tasks are sensitive to minor differences in models and parameters, the need for machine learning-based model-free methods increases as task difficulty rises. On the other hand, contact-rich tasks must be performed without damaging target objects, and the amount of training data is limited due to the nature of robotic systems. There are primarily two approaches to extracting motion models from demonstration data for skilled tasks: reinforcement learning (RL) and imitation learning (IL)~\citep{kober2010imitation,hua2021learning}. 

RL has the advantage of autonomously acquiring complex movements involving contact 
state transitions through interaction with the environment, and has been extensively studied. 
Recent trends include combining model-free and model-based RL approaches~\citep{pong2018temporal2, fan2019learning}, improving sample efficiency~\citep{shi2021proactive, wang2021deep}, pre-training in simulation followed by fine-tuning on real hardware~\citep{yang2024robot}, and integration with adaptive impedance control for contact force regulation~\citep{martin2019variable, beltran2020variable, oikawa2021reinforcement}. However, RL requires extensive trial-and-error, and learning on physical systems is often limited due to hardware wear and safety concerns. Furthermore, the complexity of contact dynamics widens the simulation-to-reality gap, making transfer more challenging. 
On the other hand, IL has the advantage of efficiently learning expert skills, including 
subtle adjustments of contact forces and positions in human dexterous manipulation. 
Expert skills inherently contain human tacit knowledge and empirical rules, and 
it is expected that if robots can acquire these in some form, their capabilities 
can be fundamentally enhanced.

Building upon the potential of IL to leverage human tacit knowledge and empirical rules, 
recent advances in Large Language Models (LLMs) have begun to extend to other modalities 
such as images and speech, and in robotics, they are expected to develop into foundational 
technologies for symbolic representation of action sequences, integration of multimodal 
knowledge, and eventually technologies that can abstract human tacit knowledge and 
empirical rules inherent in expert skills.
With the proliferation of LLMs, 
the number of papers on IL in robotics continues to increase. 
However, there is a lack of systematic organization 
of recent research trends in this field. Therefore, this paper aims to 
contribute to the advancement of robotics by conducting and systematically 
organizing a survey on IL for contact-rich tasks.

In the manipulation field, multiple surveys dealing with contact-rich tasks exist~\citep{cui2021toward, kroemer2021review, suomalainen2022survey, jiang2022review}.
An extensive review of robot learning for manipulation appears in~\citet{kroemer2021review}, providing a structured analysis of fundamental challenges, representational choices, and algorithmic frameworks. 
A systematic survey of robot manipulation examines complex interactions between robots and their environment during physical contact tasks~\citep{suomalainen2022survey}.
A thorough review of robotic assembly strategies encompasses the entire operational procedure from planning to evaluation, emphasizing the importance of integrating multiple technological approaches~\citep{jiang2022review}.

Surveys have also been published on IL~\citep{argall2009survey, fang2019survey, hua2021learning, celemin2022interactive, urain2024deep}.
A foundational survey on robot learning from demonstration in~\citet{argall2009survey} established key paradigms and methodologies that influenced the field for years.
IL in robotic manipulation was specifically addressed in~\citet{fang2019survey}, synthesizing key approaches and challenges in transferring human skills to robotic systems.
The interconnections between RL, IL, and transfer learning were explored in~\citet{hua2021learning}, highlighting their complementary nature.
A comprehensive examination of interactive IL in robotics appears in~\citet{celemin2022interactive}, emphasizing the crucial role of human-robot interaction in skill acquisition.
Most recently, the growing role of deep generative models in robotics was investigated in~\citet{urain2024deep}, particularly focusing on their application in learning from multimodal demonstrations.
No survey exists that investigates imitation learning research in contact-rich tasks. The closest paper surveyed RL technologies in contact-rich tasks~\citep{elguea2023review}. The significant differences in methodology and effects between RL technology and imitation learning make this contrast valuable for organizing trends in this field.

The contributions of this survey are as follows: 
\begin{itemize}
\item Systematic organization of existing research methods in imitation learning for contact rich tasks
\item Presentation of recent research trends and clarification of challenges
\item Presentation of future research directions and practical application possibilities
\end{itemize}

The structure of this survey is as follows. In Sec.~\ref{sec:preliminaries}, we present preliminaries from the perspectives of contact-rich robotics and motor control, and describe the key challenges in this field. Next, in Sec.~\ref{sec:collecting}, we discuss methodologies for collecting demonstrations, which are crucial for imitation learning. Sec.~\ref{sec:learningapproaches} presents trends in learning approaches, while Sec.~\ref{sec:application} showcases application cases. Finally, Sec.~\ref{sec:conclusion} provides conclusions and directions for future research.

\section{Preliminaries}\label{sec:preliminaries}
\subsection{Contact-rich robotics / Background}
In robotics, contact-rich manipulation refers to robotic manipulation tasks that involve continuous and complex interactions between the robot and its environment, often requiring sophisticated control of forces in one to several contact points. This is why direct and indirect force control (e.g., through impedance or admittance control) techniques have been widely exploited to address this problem. Pioneering contributions in this field include the formulation of operational space, hybrid and impedance control \citep{khatib1987unified, raibert1981hybrid, neville1985impedance}, as well as  advancements in contact modelling and multi-contact control \citep{whitney1982quasi,mason1986mechanics, cutkosky1989computing, bicchi1993closure}. More recent studies build upon the foundational theories proposed by these pioneering works to tackle complex contact-rich problems on advanced hardware systems \citep{ozdamar2024pushing, khandelwal2023nonprehensile}.

A major challenge in contact-rich manipulation is the scalability of classical control solutions to varying task conditions and contact dynamics. For example, even a seemingly simple assembly task, such as peg-in-hole, involves varying force or impedance requirements depending on factors like the position and orientation of the parts, their material properties, and the clearance between them. Similarly, in tasks that need continuous contact like wiping a surface or polishing, changes in surface geometry, friction, or compliance call for constant adaptation of the contact-related robot dynamics to maintain stable and effective interaction. Recent advancements in machine learning have accelerated the resolution of this problem, by enabling robots to learn intricate interaction dynamics directly from data. This shift from traditional control methods to data-driven learning has yielded significant improvements in robustness and adaptability. Techniques such as reinforcement learning, imitation learning, and adaptive model predictive control now allow robots to achieve human-like dexterity in manipulating contacts. This survey provides an in-depth analysis of these advancements.


\subsection{Insights from motor control}\label{sec:motor_control}
Humans are adaptively good but not inherently precise at contact-rich tasks. They rely on compliance, predictive control, learning, and robust feedback integration to manage contact effectively, but not on precise, optimal control like a model-based system would. This adaptive mastery has long made human motor coordination a powerful source of inspiration for robotics. Here, we provide a brief overview of key works focusing on human motor coordination in contact-rich tasks, with the aim of highlighting principles that can inform the design of more robust, adaptive robotic systems.

Motor control refers to the mechanisms by which organisms and robots orchestrate their movements to interact effectively with their environment, particularly in tasks involving significant physical contact. Unlike movements executed freely in space, contact-rich tasks require management of interaction forces, adaptation to varying material properties, and responsiveness to environmental perturbations. For instance, a robot tasked with screwing in a light bulb must precisely regulate the force applied to rotate it without damaging the glass.

Typical robotic controllers rely on predetermined mathematical models, which generally perform well in structured environments but frequently fail in real-world scenarios with uncertainty and variability \citep{kroemer2021review}. In contrast, biological motor control systems naturally manage these complexities. Humans and animals continuously adapt their movements based on sensory feedback, adjusting grip strength or modulating applied force depending on object properties and environmental conditions \citep{suomalainen2022survey,Franklin2011}. Additionally, biological systems utilize sensorimotor integration, dynamic impedance modulation, predictive control, and adaptive feedback mechanisms, enabling efficient interactions with complex environments \citep{kober2013reinforcement,Petric2017}.

Roboticists have drawn on these biological principles to develop control strategies aimed at enhancing adaptability in robotic systems. Impedance and admittance control methods, for example, manage interaction forces by dynamically adjusting the robot's stiffness and damping properties, facilitating more adaptable interactions with the environment \citep{cui2021toward, Merckaert2022}. Integrating these methods with learning-based approaches, including deep learning and reinforcement learning, has resulted in hybrid techniques that benefit from both model-based predictions and data-driven adaptability \citep{Aggarwal2022}.

Imitation learning (IL) methods particularly benefit from insights derived from biological motor control. Human demonstrations naturally encompass subtle adjustments in posture, force modulation, and adaptive responses to dynamic environmental conditions \citep{fang2019survey, Gams2022}. By incorporating principles such as dynamic impedance adaptation and sensorimotor predictive modeling, IL approaches can replicate nuanced aspects of human motor skills effectively \citep{Petric2018}.

However, challenges persist in translating biological motor control insights to robotic systems. Accurately modeling contact dynamics remains difficult due to their inherently nonlinear and sensitive nature to small variations in physical conditions \citep{jiang2022review}. Moreover, robotic sensory systems, particularly tactile and proprioceptive sensors, remain limited compared to their biological counterparts, restricting the precision and adaptability of robotic responses \citep{celemin2022interactive}. Addressing these challenges requires advancements in sensor technology and computational modeling techniques.

Future research directions will focus on improving sensorimotor integration, refining predictive control models, and enhancing impedance modulation through machine learning. Progress in wearable and tactile sensing technologies and better simulation tools for realistic contact dynamics will further enable the practical implementation of biological motor control principles in robotics \citep{hua2021learning}.






\subsection{Key Challenges}

In this section we highlight the main challenges which are currently tackled in the context of contact-rich imitation learning. 

As already mentioned, one important challenge resides in the \emph{highly nonlinear nature of contact-rich dynamics} which are extremely difficult to capture with classical modelling tools (e.g. the ordinary differential equations generated by Lagrangian mechanics) and often require computationally intractable representations (e.g. the partial differential equations generated by soft-contacts). This challenge often induced researchers to resort to learning-based model-free methods such as reinforcement learning and imitation learning. Within this context, reinforcement learning is often applied in simulation and it suffers of the intractable nature of contact models and their limited modelling accuracy (i.e. sim-to-real gap). When applied in real instead, reinforcement learning poses significant challenges for both the quantity and the nature of the required data: quantity-wise, it exposes hardware to extensive trials-and-errors which in contact-rich tasks induce a significant wear-and-tear; additionally, non-imitation learning requires explorations which by nature have the tendency to go beyond safety limits which can damage the robot and its surroundings.

On the other hand, imitation learning suffers from the  \emph{scarcity of technologies suitable for data-collection of contact-rich tasks}. Despite efforts, tactile sensors remain a technology which is mostly confined to research applications with rare industrial use-cases. Similarly, wearable tactile technologies (e.g. \citep{buscher2012}) seem to be a challenging technology to develop and their adoption haven't had yet the necessary an impact neither in research nor in industrial applications. These technologies become even more necessary since contact-rich tasks are by nature \emph{non-fully-observable} when the primary observation modality is vision. Interaction forces are an essential component in contact-rich tasks and yet they are not directly measurable from images: contact-rich tasks are often susceptible to visual occlusion since the contact area is often occluded by the body-part in contact with the manipulated object. 

Observability is even more hampered when considering the highest form of imitation, often referred to as \emph{third-person} imitation. This is the form of imitation which humans excel at: executing a task or a skill after having seen someone else performing it. As previously pointed out, first-person imitation (i.e. imitation using data of the task executed on the target robot) is in itself challenging due to the scarcity of technologies suitable for data-collection of contact-rich tasks. Third-person imitation poses additional challenges. The first challenge consists in bridging the perception gap by translating what the robot sees (third-person human actions) into its own actions (first-person robot movements) and mapping observed goals to its own perspective; this is especially challenging when relying solely on raw image data without access to complete state information. Another challenge is associated with viewpoint and appearance discrepancy: the significant differences in viewpoint (third-person human vs. first-person robot) and appearance (human arm vs. robot arm) make direct image translation or learning difficult.

Finally, in the context of contact-rich imitation the challenge of \emph{data efficiency and generalization} becomes even more challenging due to the richness of involved data and the resulting difficulties in generalizing to new tasks or objects. This is also connected to the difficulties in developing physics-based models that accurately capture and generalize the highly nonlinear nature of contact-rich interactions.

\section{Collecting demonstrations}\label{sec:collecting}
\subsection{Data modalities}\label{sec:data-modalities}

Diverse data modalities are crucial for capturing the inherent complexity of contact-rich manipulation tasks, as they enable detailed representation of both spatial and dynamic interaction properties. Position data provides essential information for precise spatial alignment and trajectory following, whereas force measurements inform the nuanced adjustments necessary for stable and compliant interactions. Vision-based modalities extend capabilities to tasks involving environmental context and indirect monitoring of contacts, and tactile information offers direct feedback on local interaction dynamics. The integration of these diverse data modalities significantly enhances the robustness and generalization of imitation learning methods in robotics~\citep{ravichandar2020recent, sherwani2020collaborative, urain2024deep}.

Positional data from joint angles and end-effector positions are fundamental for accurately replicating demonstrated trajectories. Precise positional information enables robots to achieve desired spatial configurations and smoothly transition between different motion phases. Force data complement positional information by providing necessary details about the magnitude and direction of forces applied during manipulation tasks. This information is critical for tasks requiring delicate adjustments, such as assembly or insertion operations, where appropriate force application can prevent damage to both the manipulated object and the robot itself \citep{kormushev2011imitation, peternel2015human}. 

Vision-based modalities play a significant role, particularly in imitation learning scenarios where direct interaction feedback may be limited. Visual sensors provide contextual awareness, enabling robots to interpret environmental states, object positions, and movements observed during demonstrations. However, challenges remain, such as visual occlusions and the indirect nature of force inference from visual observations. These limitations can restrict the precision of learned skills, emphasizing the necessity of complementing visual data with other sensory inputs \citep{dillmann2004teaching, vogt2017system}.

Tactile sensing and haptic feedback provide additional insights for  interactions between robotic systems and their environments, since it can enable the detection and interpretation of complex contact phenomena, such as friction, slippage, texture differentiation, and subtle deformation, essential for precise force regulation and adaptive manipulation strategies \citep{edmonds2017feeling, lambeta2024, higuera2024}. Despite significant advancements in tactile sensor design and integration methodologies, the adoption of these technologies remains primarily limited to research domain. This limitation persists due to unresolved technical challenges, which include sensor durability under repeated mechanical stress, adequate sensitivity to subtle physical interactions, and seamless integration within existing robotic platforms. Nevertheless, current developments in innovative sensor architectures, complemented by data-driven processing approaches, indicate potential enhancements in the robustness, sensitivity, and applicability of tactile and haptic feedback systems, promoting their integration in broader robotic manipulation applications \citep{edmonds2017feeling, lambeta2024, higuera2024, ablett2024multimodal}.

Effective fusion of multiple sensory modalities can further augment robotic capabilities, particularly in complex scenarios characterized by partial observability and dynamic uncertainties. Advanced multimodal integration approaches, including deep learning techniques, probabilistic inference, and filtering methods, facilitate comprehensive state estimation and enhance predictive capabilities. By leveraging complementary positional, force, visual, and tactile information, robots can achieve greater adaptability and generalization across diverse manipulation tasks. Recent research underscores the benefits of such multimodal integration, demonstrating enhanced performance and robustness in practical robotics applications \citep{urain2024deep, li2023enhancing, chen2024elemental}.

Emerging data modalities, including electromyography (EMG) signals, soft and flexible sensors, are poised to further revolutionize robotic imitation learning in contact-rich tasks. EMG has already seen applications in robotics, particularly in exoskeleton research \citep{Peternel2016}, providing insights into human muscle activation patterns, allowing robots to mimic not only observable movements but also internal force modulations \citep{Peternel2014}. Soft sensing technologies offer the potential to capture intricate interactions with complex geometries and materials, greatly enhancing sensitivity and versatility.  Recent work by \citet{liu2025forcemimic} introduces a novel hand-held device specifically designed to collect robot-free force-based demonstrations, facilitating more accessible data acquisition. Advanced wearable technologies facilitate naturalistic human demonstration capture, promising more intuitive and contextually rich data collection methodologies. The continued evolution of these modalities will likely lead to substantial advancements in the capability, accuracy, and applicability of imitation learning techniques in contact-rich robotic manipulation \citep{Zhang2022,Shih2020}.

\subsection{Teaching methods}
Robot IL is based on learning human skilled movements and is also called Learning from 
Demonstration (LfD) or Programming by Demonstration (PbD), emphasizing the aspect of learning directly 
from human demonstrations. The first process is the acquisition of human skilled movements, which shares 
many technical commonalities with teaching techniques developed in the field of industrial robots. 

\begin{figure}
  \includegraphics[width=\columnwidth]{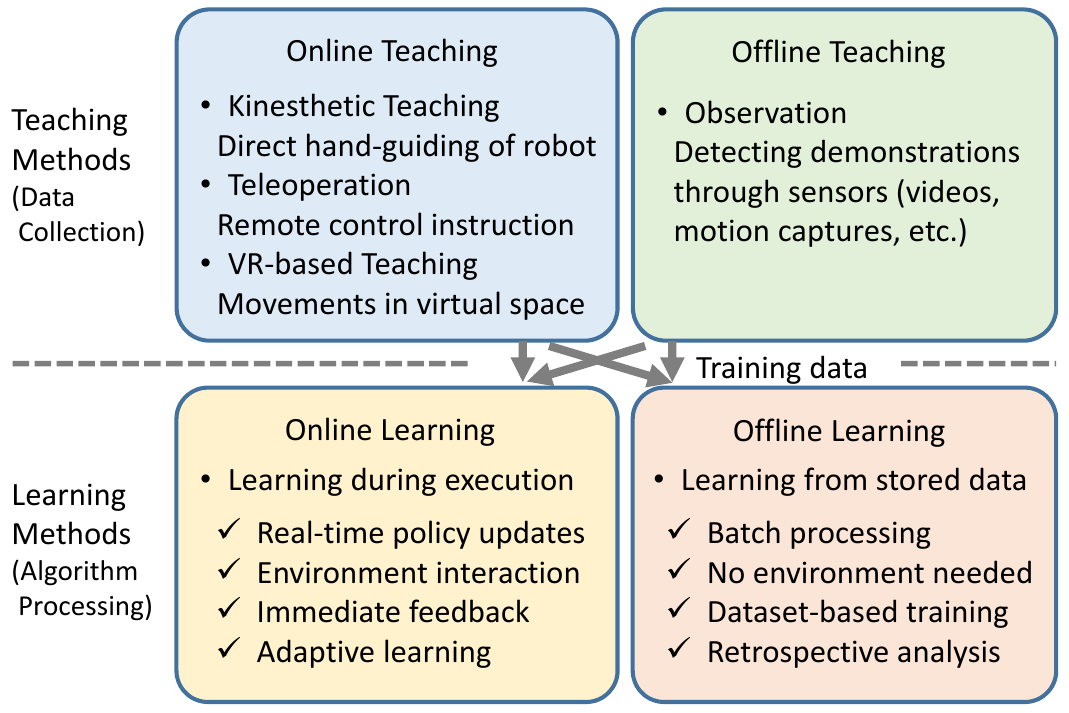}
  \caption{Relationship between robot teaching methods and learning methods}
  \label{fig:training_and_learning}    
\end{figure}

Fig.~\ref{fig:training_and_learning} shows the classification of teaching and learning. 
For imitation learning, teaching is an essential process for obtaining training data, and it 
can be divided into online teaching and offline teaching. 
The difference lies in whether trajectories are provided through online operation using a terminal, or input offline to the computer.
The concept of direct teaching is also known~\citep{ravichandar2020recent}, while it can be interpreted 
as a teaching method where the robot is directly operated online, so it is classified as online teaching. 

Online teaching is mainly classified into three categories: kinesthetic teaching, 
teleoperation, and VR-based teaching. 
Kinesthetic teaching involves direct hand-guiding of robots, while 
teleoperation utilizes remote control instructions. 
VR-based teaching, which captures movements in virtual space, is also increasingly being introduced.

In offline teaching, trajectories calculated by 
designers based on equations or trajectory-planning programs are input, or human movements are observed 
to create command trajectories. In imitation learning, only the latter method is used. 
Such methods, which detect human demonstrations through sensors, are called observation methods. 

On the other hand, machine learning is also classified into online learning and offline learning. 
It is crucial to recognize that online/offline teaching and online/offline learning are distinct concepts 
that represent different aspects to be evaluated independently in methodological frameworks.
These distinctions require careful consideration, as the terminology is frequently conflated in 
the academic literature.


Traditional industrial robot teaching assumes providing limited motions with high reproducibility 
in few trials. However, when environmental variations are large or task difficulty is high, it 
becomes necessary to acquire numerous motions and enhance adaptability through the generalization 
capabilities of machine learning. IL is effective in such cases. 
As mentioned in Sec.~\ref{sec:motor_control}, it is often necessary to acquire both 
position and force from human demonstrations to perform IL for contact-rich tasks. 
In teleoperation teaching, a leader robot operated 
by the teacher and a follower robot performing the task are connected, and bilateral control including 
force feedback is often implemented, allowing the operator's force adjustments to be transmitted to 
the follower robot~\citep{kormushev2011imitation}. 
This enables simultaneous teaching of the operator's force and position information. 
When the follower robot in teleoperation is used directly for motion 
reproduction, the reproducibility of replay is high because there is no environmental variation 
when working in the same environment. Delays in teleoperation can degrade task performance, 
while this problem is avoided by the operator's ability to compensate for those 
delays~\citep{sasagawa2020imitation}. On the other hand, in teaching through a robot, the robot's 
dynamic characteristics can interfere with skilled movements. To reduce operator burden and 
achieve high skill levels, it is essential to ensure transparency in bilateral control to minimize this interference~\citep{lawrence2002stability}. 

Meanwhile, observation teaching using cameras or motion capture to mainly acquire position 
information is also a popular method~\citep{dillmann2004teaching,vogt2017system}. 
For contact-rich tasks, force sensors need to be embedded in tools~\citep{furuta2020motion} 
or tactile gloves~\citep{edmonds2017feeling} to acquire force information. 
Methods that simulate force information during operation through the introduction
of virtual reality technology are also useful~\citep{aleotti2003toward, zhang2018deep}. 
Additionally, the combination of position and force observations with muscle activity measurement 
enables learning not only the motion but also stiffness behavior~\citep{peternel2017method}. 
Observation teaching suppresses the degradation 
of teaching due to robot interference, so the quality of preserved skilled movements is high. 
However, since dynamics differ between humans and robots, variations in environments involving 
robots are unavoidable. The robot needs to suppress these variations through some method.

Fig.~\ref{fig:training_and_learning} shows that IL can be categorized into the following four 
categories based on the combinations of online/offline teaching and online/offline learning:
\begin{itemize}
   \item Interactive Imitation
   \item Demo-Augmented Reinforcement Learning
   \item Direct Imitation
   \item Observational Learning
\end{itemize}

First, the combination of online teaching and online learning is called interactive imitation. 
The most prominent example is DAgger~\citep{pmlr-v164-hoque22a, 9551469}, 
which implements a mechanism where humans correct mistakes in real-time. 
Second, the combination of offline teaching and online learning is called demo-augmented reinforcement learning. 
GAIL~\citep{xiang2024sc, li2021meta, gubbi2020imitation,tsurumine2019generative} 
performs adversarial matching with expert demonstrations. 
DDPGfD~\citep{vecerik2017leveraging} and DAPG~\citep{rajeswaran2017learning} can also be considered as types of imitation learning since reinforcement learning is initialized with demonstrations. 
Third, the combination of online teaching and offline learning is called direct imitation. 
This category includes BC and DMP methods that use online teaching approaches such as kinesthetic teaching, teleoperation, and VR-based teaching. Finally, the combination of offline teaching and offline learning is called observational learning. 
BC from datasets involves learning from large-scale demonstration collections, 
while video-based imitation learning realizes learning from video observations.

\subsection{Synthetic data generation}\label{sec:synthetic-data}
In modern robotics, synthetic data generation has become a key enabler for the development of data-intensive learning based systems. It offers diverse and scalable environments that can be crucial for algorithms' training and validation. When it comes to robotic tasks with physical contacts, the role of synthetic data generation to enable autonomous robotic behaviors become even more significant due to the complexity, variability, and safety-critical nature of contact dynamics, which are difficult to capture, annotate, and scale in real-world data collection. Hence, synthetic data generation offers numerous advantages such as access to contact dynamics like forces and friction properties, fast domain adaptation, and risk mitigation in safety-critical scenarios such as industry and healthcare \citep{james2020rlbench,yu2020meta,mees2022calvin,liu2023libero}.

Over the past years, several platforms offering synthetic data generation in contact manipulation have been introduced. Physical simulators such as MuJoCo (Multi-Joint dynamics with Contact) \citep{todorov2012mujoco}, PyBullet \citep{coumans2016pybullet}, Isaac Gym \citep{makoviychuk2021isaac}, and Drake \citep{drake} are the most commonly used examples in the community, although every year new platforms with even stronger physics simulation capabilities emerge. MuJoCo prioritizes efficiency and smooth contact dynamics, making it a leading choice for model-based control and reinforcement learning in continuous action spaces. PyBullet, known for its user-friendly interface and extensive community support, excels in accessibility and robotic manipulation, although with less precise contact modeling. Isaac Gym, exploiting NVIDIA's GPU acceleration, enables high-throughput parallel simulations, ideal for large-scale reinforcement learning, especially in tasks involving complex contact. Finally, Drake employs hydroelastic contact mechanics, providing a principled and accurate approach, crucial for planning, control, and formal verification in safety-critical or high-fidelity applications. These simulators collectively offer a spectrum of trade-offs between computational speed, physical realism, and implementation complexity for synthetic data generation.

Regardless of the chosen simulation environment, several practical techniques have been introduced for synthetic data generation. Domain randomization \citep{tobin2017domain}, for example, is used to vary certain parameters of interest (e.g., object weight, surface texture) to make the learned models more robust to real-world variations. This method helps algorithms account for such changes and achieve performance in real-world applications that more closely matches their performance in simulation. Another method to  bridge the sim-to-real gap is based on physics-based augmentation of the generated data \citep{yang2025physics}. It enhances synthetic datasets by incorporating physical models or constraints into the data creation or transformation process. 

Despite fundamental progress in these techniques, transferring learned policies or models to real-world robots remains a significant challenge due to the sim-to-real gap, which enlarges as the number and complexity of contact dynamics increase \citep{tobin2017domain, peng2018sim, james2019sim}. This is why several recent studies have focused on fine-tuning the learned algorithms with real-world data \citep{finn2017one, yu2018one, chebotar2019closing}. In parallel, more advanced and emerging strategies, such as differentiable simulation \citep{degrave2019differentiable, brax2021}, synthetic tactile data generation \citep{9697425}, and learning from hybrid (simulated and real) datasets \citep{ferguson2020hybrid}, are gaining increasing attention for their potential to further reduce the sim-to-real gap in contact-rich manipulation tasks.

\subsection{Available Datasets for contact rich tasks}\label{sec:datasets}

The field of contact-rich imitation learning heavily relies on diverse and extensive datasets that capture interactions between robots and their environment through various sensory modalities. Several notable datasets have been developed to address the challenges of data scarcity, generalization, and multimodal integration in this domain.

The TVL Dataset \citep{fu2024} comprises 44,000 in-the-wild vision-touch pairs, featuring tactile data from DIGIT sensors and visual observations. A significant portion (90\%) of its English language labels are pseudo-labeled by GPT-4V, while 10\% are human-annotated, aiming to bridge the gap in integrating touch into multimodal generative language models. Similarly, the Touch100k Dataset \citep{cheng2024} focuses on GelSight sensors, offering over 100,000 paired touch-language-vision entries with multi-granularity linguistic descriptions. This dataset, curated from existing tactile datasets like TAG and VisGel, utilizes GPT-4V for generating detailed textual descriptions, which are then refined through a multi-step quality enhancement process involving Gemini 2 for consistency assessment. It aims to improve tactile representation learning for tasks such as material property identification and robot grasping. The VisGel Dataset \citep{8953737} also provides a large collection of 3 million synchronized visual and tactile images from 12,000 touches on 195 diverse objects, collected using KUKA LBR iiwa robotic arms equipped with GelSight sensors and webcams, to explore cross-modal prediction between vision and touch.

For broader robotic manipulation, BridgeData V2 \citep{pmlr-v229-walke23a} is a large-scale dataset with over 60,000 trajectories collected across 24 environments using a low-cost WidowX 250 robot arm. It integrates RGB images, depth data, and natural language instructions, supporting open-vocabulary task specification for various imitation learning and offline reinforcement learning methods, with a focus on generalizing skills across different environments and institutions. The DROID (Distributed Robot Interaction Dataset \citep{Khazatsky-RSS-24}) further extends this with an unprecedented scale, featuring 76,000 demonstration trajectories (350 hours of interaction) across 564 scenes, 52 buildings, and 86 tasks. Collected by a distributed network of 50 data collectors across 18 labs worldwide on the Franka Panda robot arm, DROID includes synchronized RGB camera streams, depth information, and language annotations, aiming to enhance policy performance and robustness in “in-the-wild” scenarios.

The Open X-Embodiment (OXE) Dataset \citep{openx2024} is a significant aggregation, combining over 1 million real robot trajectories from 60 existing datasets across 22 robot embodiments and 21 institutions. This large-scale repository provides diverse robot behaviors, embodiments, and environments in a standardized format, facilitating research into X-embodiment training for generalizable robot policies, including those like RT-X models. The Sparsh project \citep{higuera2024} introduces a curated dataset of approximately 661,000 images from various vision-based tactile sensors (DIGIT, GelSight) for self-supervised learning, alongside the TacBench benchmark. TacBench offers six touch-centric tasks with labeled data for evaluating touch representations, including force estimation, slip detection, pose estimation, grasp stability (from the Feeling of Success dataset), and textile recognition, demonstrating the value of pre-trained touch representations for contact-rich manipulation.

Lastly, a dataset was collected for a study on Multimodal and Force-Matched Imitation Learning with a See-Through Visuotactile Sensor \citep{ablett2024multimodal}. This dataset, created through kinesthetic teaching with a 7-DOF robotic system, includes visuotactile, wrist camera, and relative end-effector pose data, with a focus on improving imitation learning for door-opening tasks through tactile force matching and learned mode switching. These datasets collectively advance the capabilities of robots in handling complex, contact-rich tasks through multimodal sensory integration and scalable learning approaches.


\section{Learning Approaches}\label{sec:learningapproaches}
\subsection{Behavior cloning}

The substantial advancements in large-scale robotics datasets~\citep{pmlr-v229-walke23a} has led to remarkable progress in imitation learning, which leverages pre-collected data for training purposes. Notable examples include RT-1~\citep{Brohan-RSS-23}, which was trained on a dataset collected over 17 months using a fleet of 13 robots, and RT-2~\citep{pmlr-v229-zitkovich23a}, which employs vision-language models trained on Internet-scale data. These models have demonstrated impressive generalization capabilities in executing various real-world tasks. This progress has further spurred the development of large-scale robotics datasets, such as the Open X-Embodiment dataset~\citep{openx2024}. Behavior Cloning (BC) has emerged as a particularly influential paradigm, attracting substantial research interest due to its straightforward implementation and empirical effectiveness. In particular, its ability to efficiently learn from pre-collected data is especially attractive for imitation learning research.

BC constitutes a supervised learning paradigm in which an artificial agent is trained to replicate expert behavior using demonstration data~\citep{ross2010efficient, levine2016end}. Specifically, this methodology involves training the agent to generate appropriate actions $a_t$ in response to corresponding state inputs $s_t$, learning the mapping between environmental states and expert-demonstrated actions from the data distribution $\mathcal{D}$. The objective function for BC can be expressed as
\begin{equation}
J(\theta) = \mathbb{E}{(s_t, a_t) \sim \mathcal{D}} \left[ \| \pi_\theta(s_t) - a_t \|^2 \right],
\end{equation}
where policy $\pi_\theta$ is parameterized by a vector $\theta$.
This supervised learning approach is well-suited for acquiring motor skills when expert data are available, such as demonstrations corrected via teleoperation by human operators. For instance, leveraging large-scale datasets of state-action pairs has enabled robots to learn and execute manipulation tasks, such as pick-and-place operations involving objects within kitchen drawers across various domains~\citep{Brohan-RSS-23, pmlr-v229-zitkovich23a}. These advancements demonstrate the feasibility of learning intricate manipulation sequences that require precise spatial awareness and robust control strategies.

In BC, models capable of processing sequential data have been widely adopted. Representative approaches include recurrent neural network (RNN)~\citep{ito2022efficient} and long short-term memory networks (LSTM)~\citep{kutsuzawa2018sequence, rahmatizadeh2018vision, adachi2018imitation, scherzinger2019contact, funabashi2020variable}. For instance, LSTMs have been successfully applied to tasks involving deformable objects, such as closing the zipper of a bag~\citep{ichiwara2022contact}.
\citet{yang2016repeatable} demonstrated a method for teaching a robot to fold fabric using a time-delay neural network (TDNN) combined with a deep convolutional autoencoder (DCAE).
Sequence-to-Sequence (Seq2Seq) models have been utilized for learning contact-intensive manipulation tasks. \citet{kutsuzawa2018sequence} incorporated a contact dynamics model into a Seq2Seq framework with an embedded LSTM, allowing a robot to scoop and rotate objects using a spatula. Similar models have also been applied to tasks such as toilet cleaning~\citep{9223341}, where precise positional correction is required, and door opening~\citep{10161145}, which demands effective force regulation.

Transformers~\citep{vaswani2017attention} have become central to robot imitation learning, offering better handling of long sequences than RNNs or LSTMs~\citep{sherstinsky2020fundamentals} via attention mechanisms. By Action Chunking Transformer (ACT), complex tasks like opening a cup lid or putting on shoes can be learned. \citet{Zhao-RSS-23} introduced ALOHA, a low-cost teleoperation system enabling high-quality demonstrations for tasks like turning book pages or disassembling pens. With 50 demonstrations collected via ALOHA, training ACT enabled successful imitation of complex tasks. As imitation learning relies on high-quality data, advances in hardware tools like mobile ALOHA~\citep{fu2024mobile} will further drive progress.

Beyond Transformers, other generative models have been applied to BC. Diffusion models, which generate force and position trajectories via denoising, have shown success~\citep{liu2025forcemimic}. The Mamba model~\citep{jia2024mail}, a state-space model capable of handling long sequences with greater efficiency, improves generalization under limited data by focusing on salient features. It functions effectively as a motion encoder that compresses sequential robotic motion data while preserving essential temporal dynamics for accurate prediction~\citep{tsuji2025mamba}. Additionally, implicit Behavior Cloning~\citep{pmlr-v164-florence22a}, using energy-based models, offers advantages for tasks with discontinuous transitions, such as contact-rich manipulation.


A key limitation of current BC approaches is the lack of adaptability. In contact-rich tasks, when predicted actions fail to achieve desired contact states, systems must autonomously adjust their behavior. Effective adaptation requires feedback mechanisms that inform how to correct actions in response to environmental changes, ensuring robust performance in dynamic settings. In addition, a common issue in imitation learning is compounding error—the accumulation of incorrect actions over time, which can hinder proper task execution. This problem is amplified in techniques like action chunking, where action sequences are generated in blocks, increasing the risk of error propagation.

Several studies have developed the ability for autonomous behavior correction in the context of imitation learning. \citet{ankile2024imitation} has incorporated residual RL policy into base policy trained in BC manner to produce chunked actions. This approach aims to have higher frequency closed-loop control in order to correct behavior in the context of assembly task. Another approach is called Corrective Labels for Imitation Learning (CCIL)~\citep{10801414} generates some state-action pair to bring the agent back to the expert state to deal with compounding error.

To improve policy learning in BC, some studies incorporate human feedback. Language-conditioned methods guide robots with verbal cues (e.g., “move to the right”) to resume tasks autonomously~\citep{pmlr-v164-jang22a,Shi-RSS-24}. Additionally, DAgger variants like ThriftyDagger~\citep{pmlr-v164-hoque22a} and LazyDAgger~\citep{9551469} iteratively integra
te human corrections during execution, helping reduce distributional shifts and improve generalization without constant supervision.
Currently, refining policies using human language feedback has mainly explored positional trajectory adjustments.

\subsection{Dynamic movement primitives}\label{sec:dmps}

Dynamic movement primitives (DMPs) are a widely known motion representation method that facilitates generalizability and encourage convergence of the learned trajectories thanks to their dynamical system-based formulation \citep{schaal2003control}. 
The classical DMP formulation fits the target trajectory by learning the weights $w_i$ of phase-shifted Gaussian basis functions $\phi_i$, known as the \textit{forcing term}
\begin{equation}\label{eq:gauss-basis}
    f(x) = \frac{\sum{\phi_i w_i}}{\sum\phi_i}x, \phi_i=e^{-h_i(x-c_i)^2},
\end{equation}
where $c_i$ is the centre and $h_i$ is the width of each basis function. $x$ is the phase variable that decays with constant rate $\alpha_x$ towards zero in time, described as the \textit{canonical system}
\begin{equation}
    \dot{x}=-\alpha_x x.
\end{equation}
The \textit{transformation system} is formed as a spring-damper system with stiffness $\beta_y$ and damping $\alpha_y$, that drives the system state $y$ towards the goal state $g$ 
\begin{equation}
    \ddot{y}=\alpha_y(\beta_y(g-y)-\dot{y})+f(x).
\end{equation}
In the original approach, a single DMP was fit to a single \mbox{one-dimensional} trajectory of the position or joint angle.
Learning multi-dimensional motions was possible by learning multiple DMPs, each of which describing an independent motion dimension, modulated with the same canonical system.
However, this brought the limitation of representing coupled spaces such as quaternions and rotation matrices. 
Thus, some of the works focused on representing different spaces and manifolds with DMPs. 
\citet{ude2014orientation} extended the DMPs to represent quaternions and rotation matrices for non-minimal, singularity free orientation handling.
\citet{abu2020geometry} proposed the geometry-aware DMPs (\GDMP s) to support symmetric positive definite matrices that enable learning the stiffness, damping and manipulability ellipsoids as part of the motion. 
They later generalized the {\GDMP} formulation to support any Riemannian manifold including the quaternions and multi-dimensional rotation matrices \citep{abu2024unified}.

Another limitation of the classical DMPs has been capturing the variance in multiple demonstrated examples.
Probabilistic movement primitives (ProMPs) \citep{paraschos2013probabilistic} are a widely-adopted DMP-derivative that learns a probabilistic distribution over multiple trajectories. ProMPs can model the  variance in the demonstrations and combine multiple learned primitives smoothly. 
Another DMP derivative framework is kernelized movement primitives (KMPs) \citep{huang2019kernelized} that replaces the basis functions with a kernel-based non-parametric approach. KMPs have the advantages of efficiently handling high-dimensional data and allowing via-point based motion modulation. 
Probabilistic DMPs (ProDMP) \citep{li2023prodmp} unify the DMPs and ProMPs to retain the useful properties of both the dynamical systems and the statistical distributions.

Some of these works \citep{ude2014orientation,paraschos2013probabilistic,huang2019kernelized,li2023prodmp} are not contact-rich applications, but they lay the foundation for some of the contact-rich work we mention later.
In the following, we limit our discussion primarily to the recent works focusing on contact-rich imitation learning, rather than presenting an exhaustive list of movement primitive (MP) works. 
We refer the reader to an earlier survey \citep{saveriano2023dynamic} for more general and historical views on the MP literature. 
However, we also include other MP works such as those based on Gaussian mixture models (GMMs) \citep{calinon2010learning} instead of the Gaussian basis functions~\eqref{eq:gauss-basis}. These works do not directly extend DMPs, however, they were historically developed in parallel and aimed to answer similar problems with similar principles \citep{,khansari2011learning,ureche2015task}. 

As discussed earlier, the traditional DMP formulation is limited to a single modality and a single task, that is often a position trajectory in joint or task space. 
However, contact-rich tasks require high-level of context awareness that is possible with multi-modality. 
Consequently, the MP-based contact-rich manipulation methods answer this problem by either \textit{parallelization} of the other modalities as separate perception or action modules, or \textit{reformulation} of the MPs.

The modality \textit{parallelization} strategy can be traced back to early MP research \citep{nemec2013transfer,kober2015learning}. 
For example, an early peg-in-a-hole application \citep{nemec2013transfer} recorded a force profile alongside the DMPs. The DMPs were adapted according to an admittance control law to match the force profile.
Focusing on more recent works, \citet{cho2020learning} train hidden Markov model (HMM) and DMP models in parallel for each motor skill: former to select which MP to apply based on the reaction force/moment signal, latter to encode the position-based motion trajectories. 
\citet{chang2022impedance} train separate DMPs for position and force trajectories. Then, the impedance is adapted during execution to balance between position and force tracking. 
\citet{zhao2022hybrid} train GMMs for position, velocity and force profiles. Then, they adapt the impedance parameters through online optimization to achieve learned profiles.
\citet{escarabajal2023imitation} accommodate the force profile in parallel using GMMs, while using DMPs for the position trajectory.
\citet{yang2018robot} use GMMs for learning from multiple demonstrations and train a neural network-based controller to compensate for the dynamic effects.

Through \textit{reformulation}, compliant movement primitives (CMP) \citep{denivsa2015learning} add the joint torque modality to model the task-specific dynamics. \citet{Petric2018} improve the CMP framework for safe and autonomous learning of the joint torque profiles. 
Another reformulation, Bayesian interaction primitives (BIPs) \citep{campbell2019probabilistic}, integrate human monitoring modalities to achieve coordination in human--robot interaction (HRI) tasks.
\citet{stepputtis2022system} use the BIP policies to modulate the temporal progress of a bimanual multipoint insertion task. They use multi-modal sensing (force, proprioception, object tracking) to identify the task phase, and control two robot arms accordingly.
\citet{ugur2020compliant} propose the compliant parametric DMPs that learn haptic feedback trajectories through parametric HMMs and reproduce the desired force profile through a compliance control term.
\citet{qian2025hierarchical} propose the hierarchical KMPs to generalize a learned motion from known subregions to novel subregions using the correlations between the human and robot positions in object hand-over tasks.
\citet{lodige2024use} extend the ProDMP framework to be force aware (FA-ProDMP). FA-ProDMP learns the force-position correlations from multiple demonstrations to solve contact-rich tasks like peg-in-hole.

MP \textit{reformulation} does not only aim to support multi-modality, but also to answer task-specific requirements. 
\citet{yang2022learning} choose rhythmic DMPs to represent robot policies in periodic household tasks such as table wiping, food stirring and cable wiring. Unlike the classic DMP approach, the robot learns the task through visual keypoints extracted from human video demonstrations. 
Rhythmic DMPs facilitate reproducing and adjusting periodic actions. 
\citet{sidiropoulos2021reversible} propose the reversible DMPs that support backwards reproduction of a learned trajectory, which is a desirable feature to recover from errors in physical interaction and operate in unpredictable environments such as cluttered areas.
\citet{escarabajal2023imitation} use the reversible DMPs to encode the trajectory so that the mechanism can safely retract its action when assisting an injured person. 
Mesh DMPs \citep{vedove2024meshdmp} extend the \GDMP s for reproducing learned motions on complex mesh surfaces to achieve contact-rich tasks like surface polishing.

MPs provide a structural basis for a generalizable motion. The structure confines the parameter search space more than an unstructured Multi-Layer Perceptron (MLP), leading to a better sample efficiency. Thus, movement primitives are often used as the policy to be initialized using IL, and further improved  using RL. 
\citet{cho2020learning}, who train parallel HMM and DMP models, apply RL to further improve  both DMP-HMM parameters. In the case that HMMs do not identify a matching skill, a new DMP-HMM pair is learned using RL.
\citet{davchev2022residual} combine a DMP policy with RL by learning a residual correction policy to account for the contact-rich aspect of physical insertion tasks. They introduce an additive coupling term in the DMP formulation and learn this term as a nonlinear RL-based strategy. They show the advantages of perturbing the DMP policy directly in the task-space on both task success and sample efficiency. 
\citet{zang2024human} combine IL and RL by first learning ProMP policies from demonstrations and then guiding the RL training based on ProMP priors. ProMPs are chosen for their flexibility, smoothness and generalization properties.

Just like embedding or coupling another model into the DMP framework, DMPs can also be embedded into more expressive models to leverage the benefits of both. 
\citet{bahl2020neural} embed a DMP-based dynamic system into a neural network to take the advantages of both  the generalization capacity of neural networks and the efficiency of dynamical systems.
In their architecture, deep neural layers learn the parameters of DMP systems for various raw input from vision or other sensors. The parametrized DMPs then derive the motion trajectories to be executed. 
Conditional neural movement primitives (CNMPs) \citep{seker2019conditional} use conditional neural processes (CNPs) to learn sensorimotor distributions of multiple modalities. The learned CNPs are then conditioned to generate trajectories for new situations.

DMPs have been a central part of the robotic learning from demonstration since their introduction. They are actively being extended and used for contact-rich tasks beside more recent and trending methods.  
Their generalizability, sample-efficiency and transparency are invaluable for robotic tasks where data collection is costly and risk-sensitive. 
We identified the common modes in which the DMPs are employed in recent works, such as \textit{reformulation}, \textit{parallelization} and in combination with RL. 
That being said, these modes are not mutually exclusive. They have been used together in some of the works we cited above \citep{cho2020learning,davchev2022residual,escarabajal2023imitation}. 
We expect DMPs to stay present and get integrated with novel methods like generative models in the future, thanks to their fundamental structure and various proposed extensions. 

\subsection{Generative methods}
\subsubsection{Variational AutoEncoder}
Within the domain of probabilistic modeling, auto-regressive models constitute a family of architectures that maintain both high expressivity and computational tractability. These models facilitate the decomposition of log likelihood according to the following expression:
$\log p(x) = \sum_{i} \log p_\theta(x_i | x_{<i})$. 
In lieu of direct log-likelihood optimization, an alternative approach involves the introduction of a parametric inference model $q_\phi(z|x)$ over the latent variables, enabling the optimization of a lower bound on the log-likelihood. The Variational Autoencoder (VAE) objective function takes the following form:
\begin{equation}
        \footnotesize
        \mathcal{L}(\theta, \phi) = 
        \mathbb{E}_{q_\phi(z|x)} \left[ \log p_\theta(x|z) \right] 
        - D_{KL} \left( q_\phi(z|x) \| p(z) \right)
        \leq \log p(x),
\end{equation}
\normalsize
In this formulation, $\mathbb{E}_{q_\phi(z|x)}$ signifies the expectation computed with respect to the approximate posterior distribution, while $D_{KL}$ corresponds to the Kullback-Leibler divergence measure.
Using this formulation, VAE has been used for diverse behavior learning~\citep{wang2017robust}.

VAE has been employed in imitation learning by training them to reconstruct position or torque commands~\citep{abolghasemi2019pay}. 
VAE is interpreted as a means to obtain compact representations of tactile or visual signals~\citep{10341471,9561586}. It is used for skill learning via reinforcement learning~\citep{van2016stable,cong2022reinforcement}. Language is also incorporated using the paired variational autoencoder (PVAE) model~\citep{9878160}.
\citet{10752344} employed a VAE-based model to learn contact-rich tasks like wiping, using the same architecture for simulation pre-training and real-world fine-tuning. They also generated contact-maintaining motions via force feedback in latent space, enabling adaptation to surface variations. \citet{10769883} pre-trained a VAE on a large dataset of human hand-object images, which are easier to collect than robot demonstrations. They then trained a task-specific policy using a smaller, task-focused dataset, leveraging the pre-trained decoder to generate actions.

The Conditional Variational Autoencoder (CVAE) is an extension of the VAE that incorporates conditioning variables, allowing for the generation of data that is influenced by specific conditions. In imitation learning, CVAE is utilized to modify the reconstructed behavior based on these conditioning variables, enabling more adaptive and context-aware action generation.
\citet{zhang2023visual} proposed a predictive model which is a CVAE with contrastive optimization, jointly learning visual-tactile representations and latent dynamics of deformable garments.
Another approach involves training a CVAE architecture with input torque and a task-specific parameter such as a task ID, allowing the learned model to adapt its behavior according to the given task~\citep{xu2025conditional}. A similar approach has also been applied in combination with movement primitives~\citep{noseworthy2020task}, leveraging CVAE's conditioning capabilities to adjust generated motions dynamically. By incorporating conditioning parameters into the learning process, the generated motion can be modified by altering these parameters, making CVAE particularly effective for scenarios requiring motion correction, such as adjusting actions based on contact states with objects~\citep{ren2021generalization}. Moreover, \citet{mees2022matters} handled the multi-modality of free-form imitation data by encoding demonstrations into a latent plan space using a seq2seq CVAE. Conditioning the policy on these plans allows it to focus entirely on learning uni-modal behaviors.

Encoder-decoder architectures capable of conditioning, such as CVAE, have also been employed in imitation learning using Transformer models and Mamba models. ACT~\citep{Zhao-RSS-23} and Mamba Imitation Learning (MaIL)~\citep{jia2024mail} follow a CVAE-like structure, demonstrating the effectiveness of CVAE-based architectures in imitation learning for contact-rich tasks.

\subsubsection{Foundation models}\label{sec:foundational-models}
Foundation models (FMs) have shown remarkable success in generalization and reasoning capabilities in language and vision modalities. 
Vision and language were followed by audio and navigation modalities \citep{firoozi2023foundation, kawaharazuka2024real}.
Apart from the development of transformers \citep{vaswani2017attention} and other enabling technologies, the availability of internet-scale data was a key requirement for the development of FMs. 
Readily available text and image data fuelled the generalization capability of the FMs, however, this is not the case for the robotics data.
Unlike the visuolingual data, embodied interaction modalities such as proprioception and force data are not widely publicized.
Currently available robotics datasets have sample sizes between 100K-1M examples, which are incomparable to those of the LLMs \citep{kim2024openvla}.
These limitations are followed by the additional challenges of the robotics field, such as the presence of various robotic platforms, and safe real-time execution requirements of embodied systems  \citep{firoozi2023foundation}.
These challenges are especially important in case of the contact-rich applications as the safety requirements are high and real-time contact-related data is fundamental.

Vision-Language Models (VLMs) have emerged as powerful tools in robot learning, enabling robots to interpret and act upon human instructions grounded in visual scenes.
CLIPORT~\citep{shridhar2022cliport} uses two-stream architecture: a semantic stream that encodes RGB input with a frozen CLIP ResNet50 and conditions decoder layers with language features, and a spatial stream that encodes RGB-D input and fuses laterally with the semantic stream. The output is dense pixel-wise features for predicting pick and place actions.
Manipulation of Open-World Objects (MOO)~\citep{stone2023open} enables robots to follow instructions involving unseen object categories by leveraging a pre-trained VLM to extract object information, which conditions the robot policy alongside the image and language command.
RoboFlamingo~\citep{li2024vision} also uses an existing VLM by incorporating an explicit policy head, and is fine-tuned by IL only on language-conditioned manipulation datasets.
\citep{yan2024dnact} leverage NeRF for 3D pre-training to acquire a unified semantic and geometric representation. By distilling knowledge from pretrained 2D foundation models into 3D space, the method constructs a semantically meaningful 3D representation that incorporates commonsense priors from large-scale datasets, enabling strong generalization to out-of-distribution scenarios.
\citet{duan2025aha} introduce AHA, an open-source vision-language model for detecting and explaining robotic manipulation failures using natural language. Trained with FailGen, a scalable framework that generates failure data by perturbing successful simulations, AHA generalizes well to real-world failures across diverse robots and tasks.

In robot action generation using VLMs, a common approach is to use VLMs to extract semantic features from visual and language inputs, which are then fed into a separate policy module for action prediction. Recently, end-to-end Vision-Language-Action (VLA) models~\citep{pmlr-v229-zitkovich23a} that directly map images and language instructions to robot actions have gained increasing attention.
\citet{zhao2025vlas} proposed VLAS (VLA with speech instruction), which encodes visual and speech inputs, retrieves personalized knowledge via a Voice RAG module, and uses LLaMA to generate action tokens, which are decoded into continuous robot control commands.
\citet{li2025hamster} proposed hierarchical VLA models that better leverage off-domain data for robotics by first predicting a coarse 2D trajectory with a VLM, which then guides a low-level 3D control policy for precise manipulation.
The self-corrected (SC-)-VLA framework~\citep{liu2024self} enhances manipulation robustness by combining a fast action predictor with a slow, reflective system that uses Chain-of-Thought reasoning to correct failures step by step, mimicking human-like reflection.
Robocat \citep{bousmalis2024robocat} is a generalist transformer agent that natively supports multiple robotic embodiments with different action- and state-spaces. Furthermore, Robocat can self-improve by generating new data by its own model.
\citet{driess2023palm} focus on the problem of grounding multi-modal prompts for real-world manipulation planning using LLMs. 
They propose the multi-modal sentences that embed images and state embeddings in text prompts for improved embodied intelligence.
\citet{kim2024openvla} propose an open source VLA outperforming the state-of-the-art in object manipulation tasks with smaller parameter-space. 
They employ fine-tuning techniques such as low-rank adaptation \citep{hu2022lora} and model quantization \citep{dettmers2023qlora} to efficiently adapt the model on an off-the-shelf GPU. 
\citet{hao2025tla} train a policy through tactile information to execute interactive contact and collision tasks based on VLA that outperform the traditional IL methods.

In recent years, the concept of Embodied AI/embodied LLM has been discussed for application to robotics. 
\citet{chen2025emos} investigated embodiment-aware LLM-based MAS that operates a heterogeneous multi-robot system composed of drones, legged robots, and wheeled robots with robotic arms in a multi-floor house. When given a household task, the agent needs to understand their respective robots' hardware specifications for task planning and assignment. 
\citet{zhang2025badrobot} identified safety vulnerabilities in embodied AI systems that use LLM for physical robots. They introduced Badrobot, an attack method that exploits three key weaknesses: LLM manipulation within robotic systems, misalignment between language outputs and physical actions, and hazardous behaviors from flawed world knowledge. These attacks use voice interactions to make embodied LLMs violate safety and ethical constraints, highlighting critical security risks in AI-powered physical systems.

Although these works include contact-rich tasks such as object pushing and cloth folding, these are usually solved in a quasi-static way, through position control and pick-place actions. The use of FMs in dynamic contact-rich tasks where the force needs to be regulated remains to be a challenge.

\subsubsection{Other generative methods}
Recently, in IL, generative methods have been used to model the policy or the environment dynamics by generating synthetic data or actions that mimic expert demonstrations. Beyond VAEs and foundation models, there are other notable generative approaches in IL.

Diffusion Policy \citep{chi2023diffusion} is a recent advancement generating robust and multimodal action sequences for robot control. Unlike traditional policy methods, it iteratively denoises action distributions, enabling smoother and more diverse behaviors that can handle complex, real-world tasks.
Recent studies have demonstrated its superiority over traditional methods like VAEs and GAIL in contact-rich manipulation tasks.
\citet{ankile2024juicer} use diffusion policy for automatically expanding dataset size.
\citet{Prasad-RSS-24} proposed Consistency Policy, which is distilled from a pre-trained Diffusion Policy while drastically reducing inference time so that can be used in resource-limited cases.

Action Chunking with Transformers (ACT)~\citep{Zhao-RSS-23} uses a Transformer-based generative model to predict “chunks” of actions (instead of single-step actions) conditioned on past states, enabling long-horizon, temporally consistent policies. It's particularly effective for robotic manipulation tasks. ACT has been actively explored in IL, particularly for robotic control tasks where long-horizon action consistency and multi-modal behavior are critical for contact-rich tasks. \citet{10637173} propose Bi-ACT model utilizing both positional and force data, enhancing the precision and adaptability of robotic tasks.

\citet{wu2024unleashing} introduced GR-1, a GPT-style transformer pre-trained on large-scale video data and fine-tuned for multi-task, language-conditioned visual robot manipulation. GR-1 takes language instructions, observation images, and robot states as input, and predicts both robot actions and future images.

In addition, \citet{chen2024elemental} proposed a generative model called Elemental, which is designed to learn from human demonstrations and generate robot actions for contact-rich tasks. Elemental leverages a combination of generative modeling techniques, including diffusion models and variational inference, to capture the underlying structure of the task space. By learning from diverse human demonstrations, Elemental can generate robot actions that closely resemble human-like behavior, enabling effective IL in complex manipulation scenarios.

\subsection{Inverse Reinforcement Learning}
Inverse Reinforcement Learning (IRL)~\citep{ng2000algorithms} also known as reward inference~\citep{kroemer2021review}, or inverse optimal control~\citep{1971articleIOC,2017articleEP,2017articleep3795}. RL has shown a promising solution to obtain an optimal policy for contact-rich tasks in recent years~\citep{elguea2023review}.
To learn the optimal policy, a suitable reward function is significant in RL method. However, in many cases, it is engineering-consuming and unfeasible to design a comprehensive reward function, especially in some complex application scenarios, such as high-dimensional physical interactions and multi-objective manipulation tasks. The natural idea is that we can infer the reward function through expert demonstrations. In this section, we introduce the IRL method.

Different from RL learning an optimal policy via a pre-defined reward function by a trial-and-error paradigm~\citep{martin2019variable,10517611}, IRL is an inverse process, which utilizes expert demonstration as the optimal policy to train a reward function that as a crucial part of RL training for generalization.
Assuming that the expert demonstration is perfect, i.e., the optimal solution under the optimal reward function and the reward function is optimized to make the
expert policy obtains a higher reward value and then uses the RL algorithm to obtain the optimal policy under the latest reward function, and iterates this process until convergence.

However, there are challenges in IRL.  Many optimal or sub-optimal policies can match the 
 demonstrations and even more reward functions that can explain an optimal policy. Therefore, to optimize the reward functions, additional experience from the trial-and-error process is needed~\citep{Sermanet-RSS-17,nair2017combining}.

Recent work by \citet{mandi2022towards} introduced a novel multi-task training method that integrates a self-attention model and a temporal contrastive module to improve task disambiguation.
\citet{zhang2021learning} learn both variable impedance policy and reward function from expert demonstrations based on IRL framework.
\citet{xu2022robot} proposed LION net, which only utilizes images as input to learning a task by RL-based control module.

\subsubsection{Adversarial Imitation Learning}
Adversarial Imitation Learning (AIL) addresses a key challenge in inverse reinforcement learning by 
resolving the ambiguity inherent in inferring reward functions from expert demonstrations. Inverse RL 
typically suffers from the problem that multiple reward functions can explain the same expert behavior. 
By leveraging an adversarial framework, AIL incorporates a discriminator that distinguishes between expert 
and agent-generated trajectories. This mechanism forces the learned policy to produce behaviors that closely 
mimic the expert, effectively providing a robust and stable reward signal. Consequently, AIL reduces the 
dependency on hand-crafted rewards and enhances the scalability of training in complex, contact-rich tasks.
AIL~\citep{ho2016generative} is a variant of IRL that leverages adversarial training to learn the reward function. AIL is based on the Generative Adversarial Networks (GANs)~\citep{goodfellow2014generative} framework, where the discriminator is trained to distinguish between expert demonstrations and generated trajectories, while the generator is trained to generate trajectories that are indistinguishable from expert demonstrations. 
The reward function is implicitly learned through an adversarial optimization process which is formulated as a minimax game, and the discriminator's output is used to define the reward signal for the policy.
AIL has been shown to be effective in learning complex manipulation tasks, such as pick-and-place operations and assembly tasks~\citep{li2023enhancing}.

\subsubsection{Generative Adversarial Imitation Learning}
Generative Adversarial Imitation Learning (GAIL)~\citep{ho2016generative} is a variant of IRL that leverages adversarial training to learn the reward function. 
GAIL is based on the Generative Adversarial Networks (GANs)~\citep{goodfellow2014generative} framework, where the discriminator is trained to distinguish between expert demonstrations and generated trajectories, 
while the generator is trained to generate trajectories that are indistinguishable from expert demonstrations. 
The reward function is then learned by optimizing the discriminator to minimize the classification error. GAIL has been shown to be effective in learning complex manipulation tasks, such as pick-and-place operations and assembly tasks~\citep{li2023enhancing}.

Recent progress in GAIL for contact-rich tasks~\citep{pmlr-v164-lee22a, xiang2024sc, li2021meta, gubbi2020imitation} has shown promising results. Specifically, \citet{tsurumine2019generative} proposed a GAIL-based approach that incorporates contact information to improve the performance of robotic manipulation tasks.
Recent advances in contact-rich manipulation have explored the use of unified frameworks for handling multiple subtasks. For example, as demonstrated in \citet{xiang2024sc}, expert demonstrations can be leveraged to train policies across different subtasks by sharing a single critic and an identical reward function. This strategy streamlines the learning process by providing consistent value feedback and simplifies reward design, ultimately promoting improved policy convergence and robustness.
\citet{li2021meta} translate human videos into practical robot demonstrations and train the meta-policy with adaptive loss based on the quality of the translated data. Robot demonstrations are not used but only human videos to train the meta-policy, facilitating data collection.
\citet{gubbi2020imitation} achieve a peg-in-hole insertion task with a $6\,\mu\text{m}$ peg-hole clearance on the Yaskawa GP8 industrial robot.

\subsection{Multi-modal IL}\label{sec:multi-modal}

Context-awareness is a key capability that distinguishes advanced robots acting in unstructured environments from the traditional robots acting in controlled environments.
It is especially important in contact-rich tasks as their success depends on timely and appropriate response to the highly dynamic contact conditions.
Future robots should be able to understand their surroundings profoundly in order to adapt their behavior and answer changing conditions of tasks. 
Context-awareness is only possible through the inclusion of diverse modalities, since  no single modality can cover the diversity of the tasks and conditions in unstructured environments.

A recent survey \citep{urain2024deep} discusses multi-modal imitation learning using deep generative models. They classify the recent works w.r.t. the types of generative models. However, the scope of this survey is different than ours as it focuses on multi-modal deep generative models and does not focus on the contact-rich tasks.
Here, we discuss the multi-modal IL methods from the perspective of contact-rich interactions.

As discussed earlier in Sec.~\ref{sec:data-modalities}, proprioceptive data constitutes the basis of robotic manipulation. Thus, it is used as the default input in the most methods, with the exceptions of vision-based end-to-end approaches \citep{levine2016end}.
Traditionally, the force modality has been a common choice in contact-rich tasks for its natural relevance \citep{siciliano1999robot}. 
In the recent works, \citet{stepputtis2022system} use the force modality in addition to the robot and object position to identify the task phase. The phase variable synchronizes the learned behavior for different modalities. 
Some DMP-based works (Sec.~\ref{sec:dmps}) learn the force profile in parallel to the motion primitives either using another DMP \citep{chang2022impedance} or a GMM model \citep{escarabajal2023imitation}. Some of them reformulate the DMP framework to include the force modality \citep{lodige2024use,qian2025hierarchical}.
\citet{osa2018online} integrate online trajectory planning with force tracking control in a surgical task (Sec.~\ref{sec:healthcare}).
\citet{liu2025forcemimic} combine force and vision as detailed later in this section.
\citet{luo2021robust} conduct a large-scale industrial task assessment of fusing force, proprioception and vision modalities.

The tactile modality can provide more advanced information about the nature of the contact, such as surface friction, shape and curvature, or a spatial array of force readings.  
\citet{george2024visuo} combine the vision modality with a tactile array with local shape information. They propose visuotactile contrastive pretraining \citep{rethmeier2023primer} for contact-rich tasks, and show that the multi-modal pretraining improves the deployment time performance, even if the tactile encoder is removed after training.
\citet{lin2024learning}  and \citet{huang3d} separately propose bimanual teleoperation systems, and study different factors affecting the visuotactile imitation learning performance.
\citet{ablett2024multimodal} propose to learn the coupling of tactile and vision modalities through IL for better handling of contact mode switching and avoiding failures due to contact slipping.

Although direct force and tactile feedback are essentially useful for contact-rich tasks, it is still possible to achieve these tasks without them. 
Impedance control \citep{neville1985impedance}  has traditionally served as the primary approach for achieving indirect force control in robotic systems. 
\citet{zhao2022hybrid} use variable impedance control, optimizing the control parameters online to match the learned force profile. 
\citet{ugur2020compliant} add a compliance term into the DMP formulation for this purpose.
\citet{solak2019learning} use impedance control to apply grasping forces to keep an object in-hand while demonstrating and reproducing learned trajectories.
Another way to modulate force is to rely on indirect sensing modalities such as vision, however, those approaches cannot provide precise force control.

The vision modality receives extensive attention in the related work both because of its capacity to capture spatial context, and its readily available methods and datasets. 
Indeed, computer vision has been the forerunner of the machine learning research. 
The language modality has also seen a huge leap lately with the development of the transformer models.
Vision and language foundational models with outstanding capacity are getting available to aid robotic research, as discussed in Sec.~\ref{sec:foundational-models}.
Consequently, we see the trend also in multi-modal contact-rich IL.
\citet{chen2024elemental} combine visual user demonstrations with natural language instructions to learn both the reward and policy functions using a VLM model.
\citet{mees2022matters} study the effects of different algorithmic and architectural decisions on  imitation learning through vision and language modalities. They evaluate various techniques on the simulation-based visuolingual manipulation benchmark CALVIN \citep{mees2022calvin}. 
\citet{xian2023chaineddiffuser} train two generative models: a transformer-based model for predicting high-level action keypoints, and a diffusion model for generating the trajectory segments between the keypoints. Both models have access to the inputs of vision, language and proprioception.
\citet{shridhar2023perceiver} add language processing and 3D voxel modality to the RLBench environments \citep{james2020rlbench} to evaluate their transformer-based behavior cloning agent.
In the tasks where it is difficult to model the manipulated object, such as in cloth manipulation, the multi-sensory input becomes even more important. \citet{seita2020deep} propose an imitation learning system to solve fabric smoothing task through the RGB and depth modalities. 

As a special case of vision modality, offline videos have a vast potential due to the ease of collection. The video modality carries valuable info on high-level plans to solve long horizon tasks, however, it is hard to extract low-level control skills from it. This challenge is more important in the contact-rich tasks as the physical interactions are essential.
For this reason, \citet{pmlr-v229-wang23a} combine video modality and teleoperation-based proprioception modality: former for learning high-level plans, and latter for learning low-level control. They use videos of people freely playing in an environment to learn latent features about the possible actions, which are then employed to create plans to guide the low-level controller. 
\citet{iodice2022learning} use the impedance control strategy, and estimate the intended arm stiffness of the human directly from video, based on the arm configuration. They then train GMMs to learn the configuration dependent stiffness (CDS) profiles of a sawing task, and reproduce it on a robotic setup. 

Multi-modality is a key to obtain general solutions for a diverse set of scenarios.
We can categorize the works aiming to achieve diversity, or even generality into two groups: unified model and multi-model.
The former aims to develop generalist robotic policies
\citep{Ghosh-RSS-24,openx2024,bousmalis2024robocat} that aim to learn a single large model that can be easily fine-tuned to novel tasks, often employing multi-modality in both task specification, model input and action-space. 
On the other hand, the separate model approach avoids the data complexity and data heterogeneity challenges by answering the subproblems through sub-models \citep{wang2024poco,ichiwara2023modality}.
These sub-models can be combined in a hierarchical manner as 
\citet{ichiwara2023modality} proposed. Their method combines modality-specific RNN models using a high-level RNN model. \citet{wang2024poco} handles generalization by sampling the joint probabilities of separate policies as discussed below.

Diffusion models are particularly suitable for multi-modal learning due to being optimized at inference-time. This aspect enables combining multiple diffusion models when imitating the learned skills. This capacity was first shown on image generation through the composition of separately trained models \citep{liu2022compositional,nie2021controllable}.
Diffusion policies (DP) \citep{chi2023diffusion} formulate diffusion models to learn visuomotor robot policies. DPs are shown to handle the multi-modal action distributions gracefully and achieved significant improvement over the state-of-the-art methods.
Policy Composition (PoCo) method \citep{wang2024poco} combines DPs that are learned separately from many different modalities and domains, for different tasks. They achieve this by sampling the product distribution of multiple policies at the inference time. Inference time policy composition has the advantage of learning many small models, instead of a very large model like 
the generalist approaches.
Also, combined DP sampling is more straightforward in comparison to merging RNN policies \citep{wang2024robot}, which requires combined optimization of the learned model parameters.
\citet{yan2024dnact} make use of diffusion learning for combining vision, language and proprioception modalities, however, they use it for enhancing the representation learning, rather than as an action policy. 
\citet{liu2025forcemimic} adds the force modality in a DP. Their model combines the point cloud-based vision and proprioception modalities with force data.

The multi-modal approaches to contact-rich IL aim to improve task performance through the inclusion of haptic sensing, extracting knowledge from more available modalities such as vision and language, and develop generalizable solutions to large variety of tasks.
The modalities are combined in various ways such as using some modalities to select or modulate the learned skills, hierarchical use of modalities at different levels, learning joint probabilities of the modalities, extracting the task goals or costs, representation learning, pretraining, and more. 
The method depends on the used modalities and the tasks. 
The developments in the multi-modal learning depend on the collection of different sensory data, and thus, highly benefit from the availability of public multi-modal datasets and benchmarks (Sec.~\ref{sec:synthetic-data} and~\ref{sec:datasets}).
These are also useful to evaluate and compare the diverse set of proposed methods.

\subsection{Offline Reinforcement Learning}
Imitation learning and reinforcement learning (RL) are considered the most promising approaches for robot behavior acquisition. While imitation learning is constrained by the performance of expert demonstrations, RL offers the potential to surpass expert-level skills. This section focuses on RL methods for learning contact-rich tasks. Although current RL algorithms often require large-scale datasets, posing a significant bottleneck, recent advances in robotic dataset collection suggest that both RL and imitation learning will play a central role in achieving robust and scalable robotic behavior.
Among RL algorithms, offline RL~\citep{levine2020offline} is a prominent approach that seeks to solve tasks using previously collected datasets, without requiring additional interaction with the environment. In contrast, traditional online RL typically relies on extensive trial-and-error and continuous environment interaction to acquire new data. This requirement presents significant challenges in real-world robotic applications, where safety concerns such as hardware damage are critical.
Offline RL addresses these issues by learning solely from pre-collected data, thereby avoiding the aforementioned risks. This section focuses on the use of offline RL for acquiring contact-rich manipulation skills.

Offline RL algorithms are often employed for pre-training on diverse, pre-collected datasets, enabling more efficient learning on downstream target tasks. Several approaches have demonstrated the effectiveness of leveraging such offline pre-training.
\citet{Kumar-RSS-23} proposed Pre-Training for Robots (PTR), showing that combining diverse offline datasets with a small amount of target task data can significantly improve performance on new tasks. Their results highlight the potential of hybrid approaches that integrate broad prior experience with limited task-specific supervision.
In another line of work, \citet{10611575} improved learning efficiency by utilizing observation-only datasets that lack action and reward labels such as Ego4D~\citep{9879279} for pre-training. This approach shows that rich visual or state information alone can provide valuable priors for downstream policy learning.
These pre-training strategies are closely related to meta-reinforcement learning (Meta-RL), which aims to enable rapid adaptation to new tasks. For instance, \citet{9812312} proposed an Offline Meta-RL framework that leverages demonstration adaptation to quickly adapt to novel tasks, bridging offline pre-training and meta-learning principles.

In addition to pre-training, fine-tuning is key for adapting models to target tasks, and various strategies have been proposed within the offline RL framework. A major challenge is the sim-to-real gap, as simulation data often lacks real-world noise and variability. \citet{10161474} tackled this by using real-world data from safe executions of related tasks, enabling more robust transfer.
\citet{Zhang-RSS-23} showed that combining offline pre-training via IQL with online SAC fine-tuning improves performance on dexterous tasks like cherry-picking. However, \citet{pmlr-v229-rafailov23a} noted that naïve fine-tuning can cause distribution shifts and instability, and proposed a model-based on-policy method to mitigate this. Similarly, \citet{pmlr-v229-feng23a} introduced a fine-tuning approach using online data collection and a constraint balancing return estimates and model uncertainty, aiming for efficient yet safe real-world deployment.

Several studies have advanced learning with offline RL by exploring architectural innovations, data efficiency, and reward design.
One notable direction incorporates Q-learning into transformer-based models initially designed for imitation learning. For instance, Q-Transformer~\citep{pmlr-v229-chebotar23a} merges transformer structures with temporal-difference learning, enabling sequence modelling of actions in offline settings.
Other work has integrated goal conditioning and affordance models to provide structured priors or constraints that guide policy learning in complex environments~\citep{pmlr-v205-fang23a}.
\citet{10610177} proposed a two-stage framework that improves policy learning from limited data by decoupling the learning process, enhancing both sample efficiency and generalization.
To mitigate distributional shift, policy constraints and conservative learning methods are common, but often rely on complex approximations in continuous action spaces. \citet{pmlr-v229-luo23a} addressed this by introducing state-conditioned action quantization, offering a discrete, state-aware approximation.
Reward specification remains critical in offline RL. \citet{pmlr-v229-liu23c} proposed using a small amount of expert data to drive learning via intrinsic rewards, reducing or even eliminating the need for explicit extrinsic rewards.

\subsection{Other methods}

In addition to the common approaches discussed above, several other methods have been proposed to 
tackle unique challenges in this area, offering alternative perspectives and solutions.

World models offer a promising avenue for IL in contact-rich tasks by enabling robots to learn predictive representations of their environment~\citep{wu2023daydreamer}. These models, trained on demonstration data, can forecast the outcomes of actions, allowing robots to plan and execute complex manipulations more effectively. 
By capturing the underlying dynamics of contact interactions, world models facilitate the acquisition of robust policies that generalize well to novel situations, addressing key challenges in contact-rich imitation learning such as data scarcity and sim-to-real transfer.
~\citep{barcellona2025dream}

Zero or One-shot imitation learning: such as works~\citep{duan2017one,lazaro2019beyond,bonardi2020learning} that enable agents to generalize from minimal or even no task-specific demonstrations, leveraging meta-learning, transfer learning, or compositional reasoning to adapt quickly to unseen tasks. Unlike traditional BC or IRL, which often require extensive demonstrations, zero/one-shot IL focuses on extreme generalization, where policies must infer correct behavior from just one example (one-shot) or none at all (zero-shot) by relying on prior knowledge or task embeddings. This paradigm bridges the gap between IL and few-shot RL, offering a complementary perspective to offline RL and generative methods by prioritizing data efficiency at the expense of upfront training complexity.

IL on Riemannian Manifolds extends traditional methods to non-Euclidean spaces, where actions or states inherently lie on smooth, curved geometries. Work~\citep{zeestraten2017approach} leverages geometric priors to ensure stable and physically plausible policy learning. Manifold-aware IL avoids distortions caused by Euclidean approximations, improving performance in contact-rich tasks.

Several studies focus on learning from limited or suboptimal demonstrations, addressing scenarios where datasets may be sparse or exhibit noise, biases, or non-expert trajectories~\citep{kim2021demodice}. 
These methods relax the assumption of high-quality, large-scale data required by standard BC or IRL.
For small datasets, techniques like data augmentation~\citep{johns2021coarse}, meta-learning, or hybrid RL/IL frameworks are employed to extract robust policies despite limited supervision, often leveraging coarse-grained abstractions or hierarchical strategies to compensate for missing details. 
These approaches are critical in real-world settings in which collecting expert-quality data is expensive or impractical, though they may require careful trade-offs between generalization and fidelity to the demonstrations.

Recent advances have explored integrating unsupervised learning with imitation learning to reduce dependency on labeled demonstrations. By leveraging unlabeled data, these methods can learn useful representations and dynamics models that enhance policy learning efficiency and generalization.
Techniques like LOTUS~\citep{wan2024lotus} combine self-supervised representation learning with imitation policies, enabling agents to extract meaningful patterns from uncurated data before fine-tuning on limited demonstrations. 
This paradigm offers significant advantages in real-world scenarios where labeled demonstrations are scarce but unlabeled interaction data is abundant. The unsupervised components can learn robust feature spaces and dynamics that make subsequent IL more sample-efficient and adaptable.
While sharing some goals with offline RL and generative approaches, unsupervised IL methods uniquely focus on extracting value from completely unlabeled data. This positions them as particularly valuable for scaling IL to complex, open-ended environments

An emerging direction in IL incorporates explicit optimization objectives into the learning process, blending traditional imitation approaches with mathematical optimization techniques to create more robust and adaptable policies.
As demonstrated by \citet{okada2023learning}, these methods often formulate policy learning as a bilevel optimization problem or integrate constrained optimization layers directly into neural networks, enabling policies to satisfy physical and logical constraints while imitating demonstrations.
While sharing some conceptual ground with DMPs and model-based RL, optimization-based IL methods distinguish themselves through their formal mathematical guarantees and explicit constraint satisfaction mechanisms.

Some other methods such as learning sequential structure~\citep{tanwani2021sequential}, intervention learning \citep{korkmaz2025mile}, probabilistic activity grammars~\citep{lee2013syntactic} and skill retrival and adaptation~\citep{guo2025srsa,memmel2025strap} offer unique approaches to address specific challenges in contact-rich IL, complementing the primary techniques discussed above.

\section{Application cases}\label{sec:application}

\subsection{Industrial robots}\label{sec:industrial}
IL has emerged as a powerful framework for enabling industrial robots to acquire complex contact-rich skills efficiently from expert demonstrations. By imitating human expertise, IL reduces the need for extensive manual programming and facilitates rapid adaptation to varying tasks. 
This approach enhances precision and robustness in operations such as assembly \citep{scherzinger2019contact}, insertion~\citep{wang2022adaptive}, and pick-and-place operations~\citep{li2023enhancing}, ultimately leading to more flexible and autonomous robotic systems.
These tasks, which involve intricate interactions between the robot and its environment, pose significant challenges due to uncertainties in contact dynamics and the need for precise force and motion control. 
Industrial robots, which traditionally rely on programmed control policies, can greatly benefit from IL techniques to achieve greater adaptability and autonomy~\citep{liu2022robot}.

Industrial scenarios are the main application of contact-rich tasks. Such as
assembly (Peg-in-hole, insertion)~\citep{zhang2021learning}, deburring grinding~\citep{onstein2020deburring}, polishing and sanding~\citep{zeng2023surface}, and deformable object manipulation~\citep{salhotra2022learning}. 
Recent advances in IL have enabled industrial robots to learn complex manipulation tasks involving multiple contact transitions. For example, robots have been trained to perform pick-and-place operations with varying object shapes and sizes, demonstrating the ability to adapt to different scenarios~\citep{zhang2023learning}. IL has also been applied to assembly tasks, where robots learn to assemble components with high precision and efficiency~\citep{wang2023assembly}. 
These developments highlight IL's potential to enhance industrial robots' capabilities in contact-rich environments~\citep{abu2020variable}

Despite advancements in robotic control, several key challenges persist in these applications. These include achieving precise force regulation amid variable contact conditions, integrating real-time reflexive control with sensory feedback~\citep{zhang2025bresa, van2025tactile}, and ensuring stability while maintaining safety~\citep{hejrati2023nonlinear,zhang2025towards}. Addressing these challenges is critical for enhancing the robustness and autonomy of industrial robots in contact-rich tasks.

\subsection{Service and household robots} \label{sec:household-service}

Service and household robots have seen limited usage in commercial applications due to the challenges of physically interacting in unstructured environments with untrained end-users. 
Most of the existing commercial household robots focus on a specific task such as surface cleaning, lawn-mowing, or window cleaning \citep{zachiotis2018survey}, with the exception of Care-O-bot, which is designed to handle various tasks like tool-use and interaction with kitchen appliances using its two arms \citep{kittmann2015let}. 
Currently commercialized household robots such as the educational, entertainment, social, and toy robots do not require physical interaction \citep{zachiotis2018survey}. 
Thus, we review mainly the non-commercialized research in the following.

One of the main requirements of the service and household robotics is to interact with the end-user. 
This requirement entails multiple challenges such as the social acceptance, safe human-robot interaction and intuitive interfaces.
Although the professional service robot users may have former education, providing easy interaction modalities is still desirable \citep{gonzalez2021service}. 
For this reason, the natural language modality is fundamental for service and household robots.
Furthermore, it is desirable for the robot to understand the human physical and emotional states through non-verbal means. 
Thus, the multi-modality methods (Sec.~\ref{sec:multi-modal}) allowing language-based or image-based goal conditioning are important for the service robots. 

The second requirement of physically interacting with unstructured environments is  more relevant for our review. 
Typical contact-rich household activities include maintenance activities such as table wiping \citep{zhao2022hybrid,yang2022learning}, cloth folding and smoothing \citep{xiong2023robotube,seita2020deep,9551469}; 
interacting with the household objects such as doors \citep{ablett2024multimodal,bharadhwaj2023zero}, drawers \citep{yan2024dnact,liu2024self}, kitchen appliances \citep{mandi2022cacti,pmlr-v229-wang23a}, and tools \citep{wang2024robot,wang2024poco}; 
and kitchen tasks such as fruit or vegetable cutting \citep{liu2025forcemimic}, pouring liquids \citep{zhang2021learning}, and food stirring \citep{yang2022learning}. 
Currently, these tasks are solved using general-purpose mobile robots and manipulators in laboratory settings. 
Solving these tasks in a long-term cost-effective manner continues to be a key research goal.
Due to the abundance of challenging tasks in household environments, it remains as a major motivation for the advanced robotics research. 
Accordingly, existing datasets and benchmarks often include kitchen \citep{xiong2023robotube,li2022igibson}, tool-use \citep{wang2024robot}, or general household tasks \citep{james2020rlbench, fang2024rh20t,yu2020meta}.

Unlike industrial robots that deal with relatively constrained environments and tasks, the service robots should support a wider variety of actions. 
Thus, household robots aiming for generality may benefit from high-DoF dexterous manipulators like dual arms \citep{huang3d,Zhao-RSS-23}, multi-fingered hands \citep{shaw2024demonstrating}, or both \citep{wang2024dexcap,lin2024learning}. However, such systems are usually too costly \citep{an2025dexterous} to be made available for personal use.
Developing cost-effective dexterous manipulation systems stands as a challenge for contact-rich household applications.

Lastly, health-care robotic applications like monitoring, assisting and rehabilitating robots can be deployed in houses of the people in need, in addition to the professional places \citep{halicka2022smart,yang2020homecare}. 
Some of the potential contact-rich tasks in houses are mobility support, rehabilitation exercises, and activities of daily living, such as feeding, dressing and personal hygiene assistance \citep{yang2020homecare}.
Health-care robotic applications are covered in more detail in the next section.

\subsection{Health-care robots}\label{sec:healthcare}
One of the areas in the healthcare industry where automation is highly anticipated is surgical procedures. Given that surgery requires advanced skills and significant physical endurance, automation through robotic systems~\citep{5650301} is expected to offer substantial benefits. A clinical study has suggested that surgeons can benefit from the assistance of robotic systems in surgical procedures~\citep{nix2010prospective}.
Surgical operations involve the manipulation of soft tissues and organs, which necessitates a high level of dexterity. \citet{osa2018online} have developed a technique that mimics not only positional trajectories but also force trajectories based on expert human data. They demonstrated this approach using a dual-arm robotic system for the knot-tightening task.
In surgical procedures, both video and kinematic data are recorded for postoperative analysis, resulting in the establishment of a large repository of empirical data. With this in mind, \citet{kim2024surgical} employed imitation learning using visual and kinematic data to enable a dual-arm robotic system to robustly perform surgical tasks.
They have introduced Surgical Robot Transformer (SRT) which the ACT is integrated into the Da Vinci Research Kit (dVRK)~\citep{cui2023caveats} and successfully executed various surgical tasks, including tissue manipulation, needle handling, and knot-tying.

Another area in healthcare where automation is highly anticipated is rehabilitation. Patients recovering from conditions such as stroke~\citep{langhorne2011stroke} or Parkinson's disease~\citep{abbruzzese2016rehabilitation} require training to restore or maintain their physical functions. In most cases, treatment is administered by therapists, and due to the inherent complexity of working with human patients, automation through robotic systems presents significant challenges.

Robotic-assisted rehabilitation exercises~\citep{krebs1998robot, batson2020haptic, kato2024feedback} involving physical human-robot interaction require careful consideration due to the physical contact between the patient and the robot, making compliant behavior imperative for these tasks. \citet{escarabajal2023imitation} have developed a method for generating compliant trajectories for passive rehabilitation exercises, taking into account that previous positions along the trajectory are attainable for the patient. Their approach is based on imitation learning, encoding forces using GMR, and employing Reversible DMPs. The system enables self-paced rehabilitation exercises through back-and-forth movements along the trajectory in response to the patient's reactions.
\citet{lim2023adaptive} investigate the feasibility of using a general-purpose collaborative robot for rehabilitation therapies. Imitation learning methods were employed to replicate expert-provided training trajectories that can adapt to the subject's capabilities, facilitating in-home rehabilitation training.
Their approach incorporates the concept of HG-DAgger~\citep{8793698}, allowing human intervention to ensure that patients do not attempt trajectories that may be difficult to execute. By integrating imitation learning with a system that permits human intervention, this method is expected to be beneficial for in-home rehabilitation.

\section{Conclusion}\label{sec:conclusion}
This paper comprehensively surveys research trends in IL for contact-rich tasks. Contact-rich tasks, requiring complex physical interactions with the environment, represent a central challenge in robotics. The nonlinearity and complexity of these contact tasks make it difficult to address them with conventional control methods alone. Understanding everyday physics has long been recognized as a challenging problem, and this is particularly evident in contact-rich tasks where slight positional deviations can cause significant behavioral changes. Against this background, IL approaches that learn from human demonstrations have attracted considerable attention.

The paper analyzes in detail the main approaches to IL, including BC, DMP, Generative methods, and IRL. Recent developments highlight how the latest generative models have been applied to IL, demonstrating excellent performance even in complex tasks. Each method has its distinct strengths, requiring appropriate selection based on the nature of the task and available data. In response to these advances in learning technology, implementations are progressing in various fields, particularly in industrial robots (assembly and picking tasks), household robots (everyday manipulation tasks), and medical robots (surgical and physical therapy support).

Looking forward, the field demands development of adaptive control strategies, efficient learning methods from limited training data, safe and robust operation in real-world environments, and enhanced human-robot collaboration. To address these global challenges, three critical research directions emerge: 1. Design theory for dual-process hierarchical architectures, 2. Multimodal sensing, and 3. Bridging the gap between simulation and the real world.

First, designing hierarchical architectures based on dual-process theory is a crucial challenge. 
Researchers are applying the cooperation mechanism between System 1 (automatic, unconscious reactions) and System 2 (conscious, deliberative thinking) from human cognition to robotic systems. This dual-process approach aims to achieve both adaptability to complex environments and advanced decision-making capabilities.
Therefore, hierarchical architectures hold the potential to dramatically improve robot autonomy and flexibility, especially in complex tasks involving contact. While many researchers have proposed various models, a unified design principle has not yet been established. As LLMs continue to develop, expectations for System 2 models capable of handling advanced logic are growing, while contact-rich tasks particularly require high-performance System 1 models to process environmental interactions.

Second, the success of IL requires appropriate selection and integration of data modalities. Multimodal learning that integrates multiple sensory information—not just position data but also force, vision, and tactile information—significantly contributes to improving performance in contact tasks. The development of tactile sensing, in particular, enables detection and interpretation of subtle contact phenomena, and the advancement of its hardware and software, along with integration with other sensory information, is key to achieving more delicate manipulation.

Third, the transfer from simulation to real machines (sim-to-real) remains an important challenge. Due to the difficulty of modeling contact dynamics, policies learned in simulation often fail to function well in the real world. Various approaches have been proposed to address this problem, including domain randomization, physics-based augmentation, differentiable simulation, and hybrid dataset utilization. Further development in this area is essential to overcome the challenges of limited demonstration data in contact-rich tasks.

The development of foundation models is likely to influence all three challenges mentioned above, and further research progress is anticipated. To achieve more versatile and adaptive contact-rich tasks, solutions to these technical challenges and the development of integrated approaches will be necessary.

\bibliographystyle{sageh}
\bibliography{reference.bib}
\end{document}